\pdfoutput=1

\documentclass[11pt]{article}

\usepackage[]{EMNLP2022}

\usepackage{times}
\usepackage{latexsym}
\usepackage{booktabs}
\usepackage{multirow}
\usepackage{bm}
\usepackage{url}
\newcommand{\stitle}[1]{\vspace{0.3ex} \noindent{\bf #1}}

\usepackage[T1]{fontenc}

\usepackage[utf8]{inputenc}
\usepackage{fontawesome5}

\usepackage{microtype}

\usepackage{natbib}
\usepackage{graphicx}
\usepackage{amsmath}
\usepackage{mathrsfs}
\usepackage{amsfonts}
\usepackage{amssymb}
\usepackage{booktabs}
\usepackage{multirow}
\DeclareRobustCommand{\cmss}[1]{{{\fontfamily{cmss}\selectfont{#1}}}}
\usepackage[ruled,linesnumbered, vlined]{algorithm2e} 
\usepackage{times}
\usepackage{latexsym}
\usepackage{subcaption}
\usepackage{float}
\usepackage{soul}
\usepackage[normalem]{ulem}

\usepackage{enumitem}
\usepackage[subtle]{savetrees}

\newcommand{\removelatexerror}{\let\@latex@error\@gobble}

\title{Towards Relation Extraction from Speech}

\newcommand{\monash}{\triangle}
\newcommand{\seu}{\diamondsuit}
\newcommand{\zju}{\dagger}

\author{Tongtong Wu\Thanks{~~denotes the equal contribution.}$\ \ ^{,\seu,\monash}$~ Guitao Wang$^{*,\seu}$~ Jinming Zhao$^{*,\monash}$~ Zhaoran Liu$^\zju$ \\
\textbf{Guilin Qi}$^{\seu}$~ \textbf{Yuan-Fang Li}$^{\monash}$~ \textbf{Gholamreza Haffari}$^{\monash}$ \\ [4pt]
$^{\seu}$Southeast University, China; $^{\zju}$Zhejiang University, China; $^{\monash}$Monash University, Australia \\[4pt]
$^{\seu}$\texttt{\{wutong8023,220222117,gqi\}@seu.edu.cn}, $^{\zju}$\texttt{jiuyinlau@gmail.com}\\[4pt]
$^{\monash}$\texttt{\{first\_name.last\_name\}@monash.edu}\\[4pt]
}

\begin{document}
\maketitle
\begin{abstract}

Relation extraction has focused on extracting semantic relationships between entities from the unstructured written textual data.
However, with the vast and rapidly increasing amounts of spoken data, relation extraction from speech is an important but under-explored problem. 
In this paper, we propose a new information extraction task, speech relation extraction (SpeechRE). To facilitate further research, we construct the first synthetic training datasets, as well as the first human-spoken test set with native English speakers.
%
%
We establish strong baseline performance for SpeechRE via two approaches. The pipeline approach connects a pretrained ASR module with a text-based relation extraction module. The end-to-end approach employs a cross-modal encoder-decoder architecture.
Our comprehensive experiments reveal the relative strengths and weaknesses of these approaches, and shed light on important future directions in SpeechRE research.  
We share the source code and datasets on \url{https://github.com/wutong8023/SpeechRE}.

\end{abstract}

\section{Introduction}
\label{label:intro}

Relation extraction (RE)~\cite{re_survey} is an important information extraction task, which aims at extracting structured semantic relations between entities from unstructured data, typically text. 
Besides text, there is also a plethora of speech data that is being continually produced. These include news reports, interviews, meetings and dialogues, to name a few. Extracting  relations from speech is an important but under-explored problem. 
%


\begin{figure}
    \centering
    \resizebox{.48\textwidth}{!}{
    \includegraphics{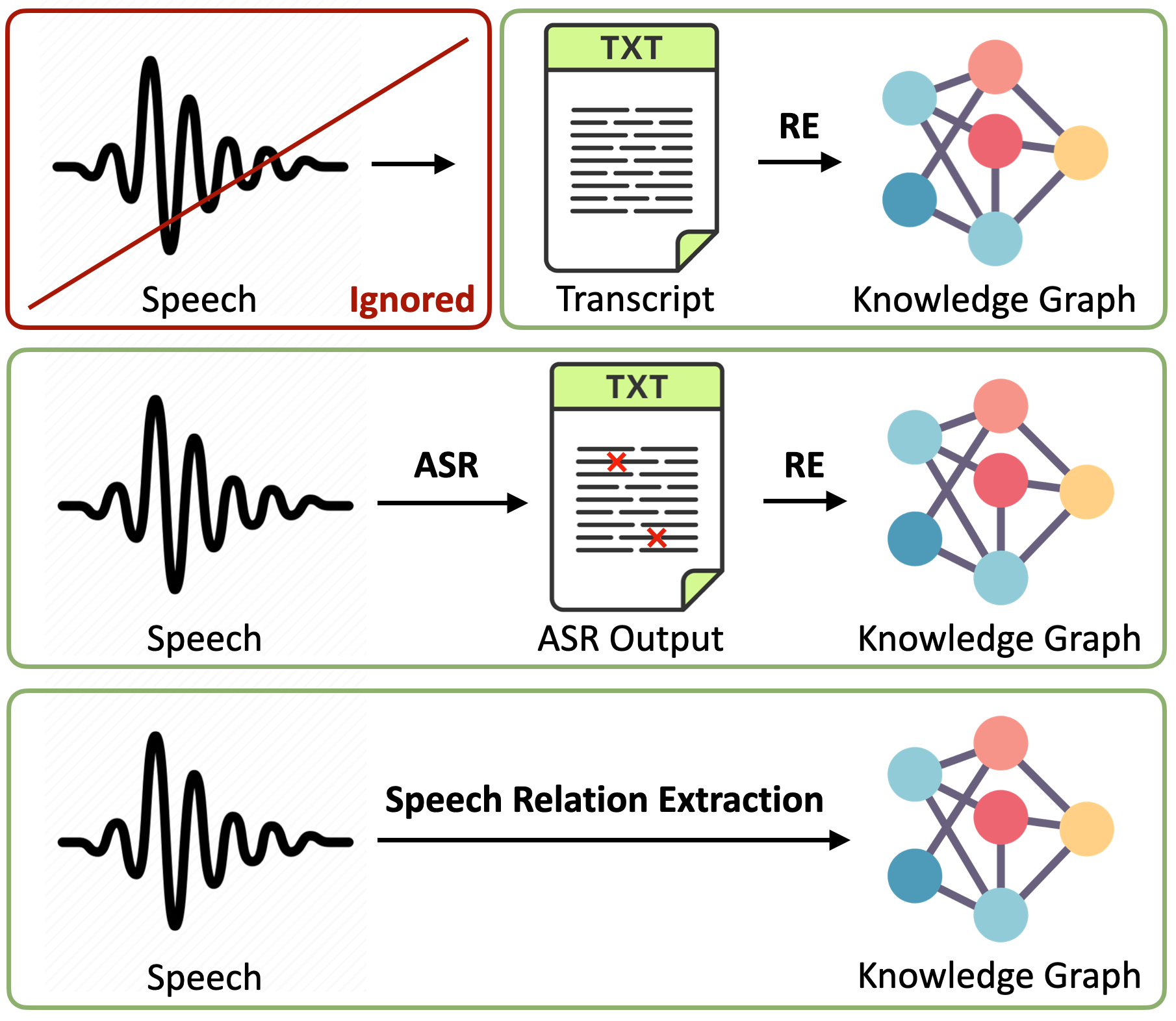}}
    \caption{The comparison among conventional transcript-based relation extraction, ASR outputs-based relation extraction, and the end-to-end speech relation extraction.}
    \label{fig:motivation}
\end{figure}

In this work, we take the first step towards addressing relation extraction (RE) from speech, introducing a new information extraction task, Speech Relation Extraction (SpeechRE). The input for this task is raw audio and the output is one or more triplets, each of which representing a relation between a pair of two entities appearing in the speech, e.g., \emph{[entity1, relation, entity2]}.
%
%

SpeechRE and text-based RE (TextRE)
both involve content understanding.
The former is more challenging than the latter, mainly due to the characteristics of speech. 
(i) Speech carries much richer information beyond linguistic content (unlike text), for instance, emotion, speaker style and background noise; and it is non-trivial to disentangle the content element~\cite{mohamed2022self}, which is needed for SpeechRE. 
(ii) Speech is continuous without sequence/word boundaries, implying the difficulty of determining the exact audio spans for target words (entity and relation). 
(iii) Audio signals are orders of magnitude longer than the corresponding transcripts, which makes speech encoding for long-span extraction more challenging due to more demanding hardware requirements, especially with Transformer \cite{vaswani2017attention}.

In the absence of SpeechRE training data, we construct three benchmark datasets for this task by converting two commonly used TextRE datasets (i.e., CoNLL04 \cite{roth2004linear} and ReTACRED \cite{stoica2021re} to speech with a SOTA text-to-speech (TTS) system. We then pair the  synthetic speech with the corresponding target relation triplets as instances. To better evaluate model performance on real speeches, we also compile a human-read test set. 



We approach SpeechRE with a pipeline method, SpeechRE$_{pipe}$, and an end-to-end (e2e) method, SpeechRE$_{e2e}$. In SpeechRE$_{pipe}$, we train our pipeline model with an automatic speech recognition (ASR) module that converts speech to text, followed by a RE module that extracts triplets. 
In SpeechRE$_{e2e}$,  we build a single speech-to-text model that extracts triplets directly from speech. We use a SOTA pretrained speech encoder, \textsc{Wav2Vec2 (W2V2)}~\cite{baevski2020wav2vec} and  BART \cite{lewis2020bart} decoder. Inspired by \citet{li2020multilingual}, we attach a length adaptor on top of the encoder to bridge the length mismatch between speech representation and text representation. 
The end-to-end approaches in speech processing tasks often face more severe data scarcity issues than the pipeline approaches \cite{sperber2020speech}, as the latter can essentially leverage massive ASR data and labelled data for the downstream text-based tasks. To tackle this challenge, we further propose two data augmentation techniques: upsampling via generating speech with different voices, and pseudo-labelling \cite{he2021generate} by leveraging abundant ASR data and a SOTA TextRE system. 

Our contributions can be summarized as follows.

\setlist[itemize]{align=parleft,left=0pt..1em}
\begin{itemize}
    \item We present a new task, Speech Relation Extraction (SpeechRE). To support the development of this task, we create and release a synthetic SpeechRE dataset, including training/dev/test sets,  as well as a human-read test set. 
    \item We establish strong baseline performance via a pipeline approach and an e2e approach. Our extensive experiments identify a performance gap between TextRE and SpeechRE, and the gap between the pipeline approach and the e2e approach, motivating further research.
    \item Our analysis shows that the performance gap of the end-to-end approach mainly comes from the data scarcity problem and the difficulty of spoken name recognition. We propose two data augmentation methods to the problem.
    \item Based on our findings, we suggest three main directions for future exploration to advance speech relation extraction.
\end{itemize}
\vspace{-0.2cm}

\section{Related Work}
\vspace{-0.2cm}
\label{sect:rela_work}
\subsection{Relation Extraction}
As an essential component of information extraction, named entity recognition (NER) and relation extraction (RE) have attracted much attention in the research community.
Relation extraction is usually studied as a natural language processing task of textual data~\cite{nasar2021named,wu2021curriculum,zheng-etal-2021-prgc,ChenZXDYTHSC22}. 
With the widespread of multimedia data on social media, some researchers have begun to explore relation extraction from data in other modalities such as images~\cite{zheng2021multimodal,HVPNET, MNRE}.
Although some work has focused on spoken language, such as dialogue relation extraction~\cite{yu2020dialogue,ZhouJL21}, these studies are all based on transcripts, i.e.\ high-quality transcribed text from speech, which is still within the confines of text-based relation extraction. Moreover, given the transcribed text, the side information of voice, e.g., emotion, speaker identity is ignored from the spoken language.
\vspace{-0.2cm}
\subsection{Spoken Language Understanding}
Spoken Language Understanding (SLU) aims to extract the meaning from speech utterances. It has wide applications from voice search to meeting summarization and has received great attention from industry and academia~\cite{tur2011spoken}. A typical SLU system involves mainly two tasks, i.e., intent detection and slot filling~\cite{tur2011spoken}. Traditionally, SLU systems have a pipeline structure, in which an ASR module is first used to convert speech to text and then a NLU system is deployed to determine semantics from text. 

A major drawback of this approach is that each module is trained and optimized independently~\cite{serdyuk2018towards}. (i) The ASR model is optimized to minimized Word Error Rate (WER)\footnote{WER is a commonly used metric in measuring the performance of ASR systems.}, often equally weighting every word, whereas not every word has the same impact on SLU. (ii) The NLU model is trained on clean text without ASR errors, i.e.\ transcripts. During evaluation, however, it receives erroneous ASR outputs and these errors are propagated to NLU, impairing its performance. End-to-end (e2e) learning has thus attracted interests from the community, for its potential to addressing SLU in a more principled way~\cite{serdyuk2018towards}. Since the first e2e approach proposed by~\citet{serdyuk2018towards}, the field has made significant advances~\cite{qinsurvey} and many techniques, such as pretraining~\cite{castellucci2019multi}, have been proposed. 

Similar to other speech processing tasks (e.g.\ speech translation~\cite{sperber2020speech}), SLU also faces the data scarcity issue, as it can be very expensive to annotate such a dataset, whereas the pipeline method can benefit from existing and emerging massive ASR data and NLU datasets. 

Speech relation extraction (SpeechRE) is a new SLU task and thus inherits the merits and demerits of the pipeline and e2e approaches. We leverage advances that have been developed in related disciplines in this work and evaluate their relative strengths and weaknesses in \S\ref{sec:exp}.

\section{Speech Relation Extraction}
\label{sect:sre_mtd}
We define SpeechRE as a joint entity and relation extraction task that takes a speech utterance as the input and generates a set of relational triplets in the form of \emph{[entity1, relation, entity2]} as the output.

In this section, we first describe the data construction method (\S\ref{data}). Next, we present our two approaches to the task (\S\ref{app}). Last, we describe our two data augmentation techniques (\S\ref{data_aug}) to improve end-to-end SpeechRE performance. 




\vspace{-0.2cm}
\subsection{Dataset Construction}\label{data}

\stitle{Synthetic Data.} As there is no readily available SpeechRE data, we generate SpeechRE data from existing TextRE corpora. Given a TextRE dataset consisting of pairs of <{source (i.e.\ transcript), triplet}>
we convert the {transcript} to human-like speech with a TTS model. A typical TTS system comprises a Text-to-Spectrogram module, which takes discrete text as input and produces mel-spectrograms, and a vocoder, which converts the mel-spectrograms  into waveforms. We choose Tacotron2-DCA as the TTS system and Mulitband-Melgan as the vocoder.\footnote{\url{https://github.com/mozilla/TTS}} 
Once the synthesis process is complete, our data would contain triples of <{synthetic speech, transcripts, triplets}>. Training/dev/test sets are compiled following this process, while obeying the original TextRE split.

\stitle{Real Data.} To evaluate the performance of our models on realistic speech, we randomly choose 200 instances from the ReTACRED10 test set 
and engage a native English speaker to read the corresponding transcripts. This real SpeechRE test set can be used as a benchmark for future research. Please refer to \S\ref{case_study} for demonstration of synthetic and real data.

\begin{figure*}[t]
    \centering
    \includegraphics[width=\textwidth]{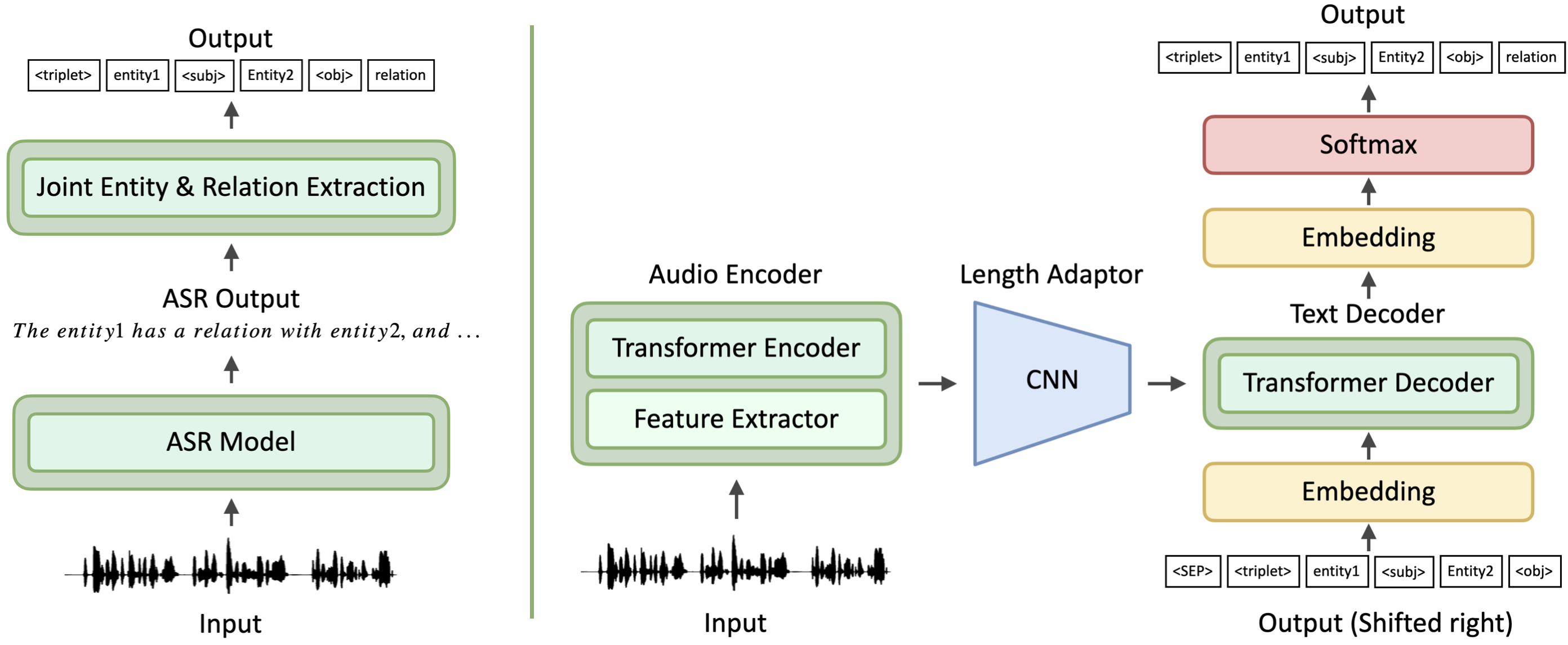}
    \caption{Overview of SpeechRE$_{pipe}$ (left) and SpeechRE$_{e2e}$ (right).}
    \label{fig:arch}
\end{figure*}

\vspace{-0.2cm}
\subsection{SpeechRE Approaches}\label{app}
We describe our pipeline (SpeechRE$_{pipe}$) and end-to-end (SpeechRE$_{e2e}$) approaches in this section. 
As depicted in Figure \ref{fig:arch}, SpeechRE$_{pipe}$ consists of an ASR module for turning speech into text and a TextRE module for extracting triplets from the text, whereas SpeechRE$_{e2e}$ has a simple architecture with a speech encoder, a length adaptor and a text decoder, which outputs triplets directly.

\stitle{The Pipeline Approach.}\label{pipe}
We use \textsc{W2V2}-large as our ASR module. It is a speech encoder, pretrained in a self-supervised manner. Its architecture starts with a feature encoder composed of several 1D convolutional neural networks that process raw waveforms and emits latent speech representations. Following that, a quantization module is attached to extract discrete latent vectors. Next, a context encoder made of 24 Transformer \cite{vaswani2017attention} layers is used to learn  contextualized representations from masked outputs of the feature encoder. The whole model is optimized to discriminate a true masked vector from the ones produced by the model. After pre-training, only the feature and context encoders are retained and used for downstream tasks. Compared with other ASR models, W2V2 obtains superior performance by fine-tuning it with a small amount of labelled speech. Additionally, it works on raw audio signals directly, avoiding the risk of information loss using hand-crafted features \cite{latif2020deep}. The \textsc{W2V2} model we use is already fine-tuned on ASR data and we do not further fine-tune it. 

%
We utilize REBEL~\cite{cabot2021rebel} as our TextRE module. It uses a pretrained language model BART~\cite{lewis2020bart} as the backbone and treats TextRE as a text generation task. Concretely, the input is text (the output from ASR in our case), and the output is linearized triplets, in the form of \emph{``<triplet> entity1 <subj> entity2 <obj> relation''}. 
The generative model may not restrict the output entities exactly the same with a mention in input text, which is an advantage for SpeechRE with ASR outputs, such a task extracting from the text containing noisy entity mentions.

An alternative to REBEL is to employ a classification-based model as our TextRE module. We experiment with Spert~\cite{spert} and TPlinker~\cite{wang2020tplinker} in this work.
%
Given that the ground-truth entity may not appear in the transcribed text computed by ASR, we modify the ground-truth entity in the training set by 
referring to the fuzzy-matched longest substring mentioned in transcribed text, such a fuzzy-matched longest substring is measured by Levinstein distance.\footnote{https://pypi.org/project/fuzzywuzzy/} 

\stitle{The end-to-end Approach}\label{e2e}
We formulate SpeechRE$_{e2e}$ as a speech-to-text task that requires a speech encoder and a text decoder. 
We employ the aforementioned \textsc{W2V2} as our encoder, for its capabilities of encoding general-purpose knowledge. We take the decoder component of BART-large~\cite{lewis2020bart} as the text decoder. Na\"ively jointing them may lead to optimization issues, as they are pretrained on different modalities which differ significantly in length. To address this issue, inspired by \citet{li2020multilingual}, we introduce a length adaptor made of $n$ number of 1-d convolutional layers, each of which is parametrized with kernel $p$, stride $s$ and padding $p$. This adaptor has an sequence reduction effect of $\sim{s^n}$. 

We follow the partial training strategy used by \citet{gallego2021end}. We train the length adaptor together with part of the encoder and decoder (including encoder self-attention, encoder-decoder cross-attention and layer normalization), while freezing the rest of the parameters. The trained parameters account for $\sim{20\%}$ of the entire model. This training strategy has shown to be efficient, while retaining performance in speech translation tasks \cite{zhao2022m}. 
\vspace{-0.3cm}
\subsection{Data Augmentation for Speech Relation Extraction}\label{data_aug}
To address the data scarcity issue facing SpeechRE$_{e2e}$, we propose two data augmentation techniques: upsampling and pseudo-labelling. For \textit{upsampling}, given a SpeechRE corpus, we use a multi-speaker TTS system~\cite{kim2021conditional} and generate synthetic speeches with 4 different voices. This yields 4 more synthetic SpeechRE datasets.
\textit{Pseudo-labelling} has been widely used in NLP \cite{he2021generate} due to its effectiveness in improving task performance. Specifically, given a SpeechRE dataset $\bm{D}$,  we fine-tune the pretrained REBEL model on <\cmss{transcript, triplet}> training instances to adapt it to the current domain. Next, we run the fine-tuned REBEL over the large-scale English dataset of CommonVoice (V9)\footnote{\url{https://commonvoice.mozilla.org/en/datasets}} where audios are recorded by volunteers of different demographic characteristics. Together with the original speech, this gives a total of 922k instances containing \cmss{<real\_speech, transcript, pseudo\_triplet>}.  
Then, we filter out noisy data if a pseudo\_triplet meets any of the following criteria: i) relation is ``no\_relation''; 2) no subject/object entity is generated; and 3) subject and object entities are both pronouns. We thus obtain 380k clean instances. Depending on the type of relations available in $\bm{D}$, further filtering may be applied to remove spurious relation triplets.







\section{Experiment}\label{sec:exp}
\subsection{Dataset}
We conduct experiments on our proposed SpeechRE datasets, i.e., Speech-CoNLL04 and Speech-ReTACRED, aligning with their original dataset CoNLL04~\cite{conll} and ReTACRED~\cite{retacred}. Furthermore, we pick 10 relations with the largest number of instances from the ReTACRED dataset, and remove the instances with none of relation or containing the other 30 relations. We named the sub-dataset of ReTACRED as ReTACRED10, and utilize it as the test-bed for sufficient supervised learning. We detail data statistics in Table~\ref{tab:data}.

Moreover, we use ReTACRED10 to fine-tune REBEL for  pseudo-labelling. The fine-tuned model generates pseudo labels  of 137 relations. We remove pseudo instances whose labels fall out of ReTACRED10. We then have 363k instances remained and each instance has one triple. Furthermore, we sample from pseudo set to 1.8$\times$ for each relation in ReTACRED10. At this point, the total number of data sampled is 2.5$\times$ to ReTACRED10, as a transcript in ReTACRED10 often has multiple relations. 



\begin{table*}[h]
\centering
    \resizebox{.9\textwidth}{!}{
    \begin{tabular}{ c | c | c | c | c | c }
    \hline
    \multirow{2}{*}{Datasets}   & \multirow{2}{*}{\# Relations} & \# Instances & \# Triplets & \# Avg. tokens  & \# Avg. audio length \\
       &     & (train || dev || test) &  (train || dev || test) & (in transcripts)  & (in seconds) \\
    \hline
    CoNLL04    &   5     &   {922 || 231 || 288}      &   {1,283 || 343 || 422}      &         29.1     &       {11.3 }\\
    ReTACRED   &   40     &   {33,477 || 9,350 || 5,805}      &    {58,465 || 19,584 || 13,418}     &      36.3        &        {12.9 }        \\
    ReTACRED10 &    10    &  {11,116 || 3,892 || 2,513}       &     {15,665 || 5,970 || 4,204}    &       34.7       &        {12.6 }        \\ 
    \hline
    \end{tabular}}
    \vspace{-0.2cm}
    \caption{Dataset statistics.}
    \label{tab:data}
    
\end{table*}


\subsection{Baselines}


We select three joint entity and relation extraction methods as baselines:
\stitle{TP-Linker}~\cite{wang2020tplinker} formulates joint extraction as a token pair linking problem and introduces a handshaking tagging scheme that aligns the boundary tokens of entity pairs under each relation type. 
\stitle{Spert}~\cite{spert} formulates the task as a two-stage classification task, with classifying each candidate continuous span for entity detection and then classifying the inter-context for relation classification.
\stitle{REBEL}~\cite{cabot2021rebel} treats joint entity and relation extraction as a text generation task.
We also attempt these three methods as the pluggable TextRE modules in SpeechRE$_{pipe}$.

\subsection{Evaluation metrics}

Evaluating SpeechRE is difficulty because of the strict matching of entities. The error of a letter or the difference in case lead to failure in entity matching, which lowers the results of triplets. For this reason, we evaluate the baseline models and SpeechRE models based on the metrics (i.e., Recall, Precision and micro-F1) commonly used in TextRE, with modifications. 
Specific to entities, we ignore the span of entities due to the lack of span information in audio, and TextRE applies the same method. 
Additionally, we do not consider entity when evaluating relations, which is equivalent to the task of relation classification of sentences. The reason is that spoken entity recognition is a very difficult tasks as most entities have low frequency in a dataset; when predicted entities were taken into account, the results would cover the true performance of relation generation. When evaluating triplets, we make sure that the head entity and tail entity and the relation between them are all correct. 






\begin{table*}[t]
\centering
\resizebox{.9\textwidth}{!}{
\begin{tabular}{@{}cc|ccc|ccc|ccc@{}}
\toprule
\multicolumn{2}{c|}{\multirow{2}{*}{Method}}                             & \multicolumn{3}{c|}{CONLL04} & \multicolumn{3}{c|}{ReTACRED} & \multicolumn{3}{c}{ReTACRED10} \\ \cmidrule(l){3-11} 
\multicolumn{2}{c|}{}                                                    & Entity  & Relation & Triplet & Entity  & Relation  & Triplet & Entity  & Relation  & Triplet  \\ \toprule
\multicolumn{1}{c|}{\multirow{3}{*}{TextRE}}                 & TP-Linker & 78.63   & 83.49    & 58.56   & 50.46   & 51.83     & 20.39   & 65.51   & 65.17     & 37.01    \\
\multicolumn{1}{c|}{}                                        & Spert     & 76.38   & 81.83    & 63.45   & 60.26   & 63.48     & 21.46   & 64.88   & 64.72     & 34.61    \\
\multicolumn{1}{c|}{}                                        & REBEL     & 85.36   & 89.86    & 71.46   & 60.09   & 65.15     & 25.15   & 64.91   & 69.80     & 39.68    \\ \midrule \midrule
\multicolumn{1}{c|}{\multirow{3}{*}{SpeechRE$_{pipe}$}} & TP-Linker$_{pipe}$ & 32.41   & 77.54    & 8.70    & 28.60   & 51.43     & 6.77    & 38.19   & 61.85     & 13.79    \\
\multicolumn{1}{c|}{}                                        & Spert$_{pipe}$     & 28.95   & 75.44    & 10.47   & 33.20   & 58.36     & 7.10    & 55.23   & 57.42     & 27.43    \\
\multicolumn{1}{c|}{}                                        & REBEL$_{pipe}$     & 35.78   & 82.86    & 12.53   & 30.21   & 53.20     & 6.93    & 51.08   & 67.46     & 28.06    \\ \midrule
\multicolumn{1}{c}{SpeechRE$_{e2e}$}                   &   & 24.89   & 59.57    & 12.50   & 27.70   & 52.10     & 6.59    & 29.87   & 51.32     & 14.79    \\ \bottomrule
\end{tabular}}
\vspace{-0.2cm}
\caption{Main results. \textbf{Upper rows}: TextRE models for which inputs are transcripts. \textbf{Middle rows}: SpeechRE$_{pipe}$ where inputs are ASR outputs. \textbf{Bottom row}: SpeechRE$_{e2e}$ where inputs are speech.} 
\label{tab:main}
\vspace{-0.2cm}
\end{table*}

\subsection{Implementation Details}
We use a pretrained \textsc{w2v2} model\footnote{\url{https://huggingface.co/facebook/wav2vec2-large-960h-lv60-self}} to convert speech to text, without fine-tuning it. Since the ASR outputs the model produces are all lower-cased without punctuation, we perform post-processing on the outputs for punctuation restoring and casing with another pretrained model.\footnote{\url{https://huggingface.co/flexudy/t5-small-wav2vec2-grammar-fixer}} For TextRE model, we mostly follow the instructions in \citet{cabot2021rebel} and start from the REBEL\footnote{\url{https://github.com/Babelscape/rebel}} that using Bart\footnote{\url{https://huggingface.co/facebook/bart-large}} as the pretrained model. The original REBEL labels the entities in the input text using punctuation marks to indicate entities’ position in the input. Since SpeechRE$_{pipe}$ does not use labels to bias the model with entity information from plain audio or text, we remove the entity labels when preprocessing ReTACRED (source sentences) and ReTACRED10. 


To train our SpeechRE$_{e2e}$ model, we use the pretrained \textsc{w2v2}  large\footnote{\url{https://dl.fbaipublicfiles.com/fairseq/wav2vec/wav2vec_vox_960h_pl.pt}} and the pretrained Bart\footnote{\url{https://dl.fbaipublicfiles.com/fairseq/models/bart.large.tar.gz}}. We keep the \textsc{w2v2} feature extractor frozen. We set kernel size, stride and padding to 3, 2, 1 for all 3 CNN layers for the length adaptor. We apply data augmentation \cite{potapczyk2019samsung} on the audio data on the fly by applying the effects of "tempo" and "pitch" to change the speech speed, and "echo" to simulate echoing in large rooms. We train our SpeechRE models for $23k$ updates and set early stopping of 20 updates. We use Adam \cite{DBLP:journals/corr/KingmaB14} optimizer with parameters (0.99, 0.98), while setting clip norm to 20. We use the learning rate to 1e-4, monitored by a tri-stage scheduler.  All experiments are conducted with fairseq.\footnote{\url{https://github.com/facebookresearch/fairseq}}. All our models are evaluated on the best performing checkpoint on the validation set. All experiments are conducted in a V100 GPU. Full training details can be found in \emph{Appendix} \ref{appendix:training details}.

\subsection{Results of TextRE and SpeechRE}
We first compare and contrast among the text relation extraction method and two speech extraction methods, to understand the performance gap. We train various models, including three TextRE models, the pipeline version of them\footnote{Where the input is changed to ASR outputs, instead of transcripts in TextRE.} and the e2e model, over CoNLL04, ReTACRED and ReTACRED10. Results are summarized in Table \ref{tab:main}.

Despite of the good performance of REBEL with transcripts being inputs (up to 71.46 on CoNLL04), its performance drops hugely when the input becomes ASR outputs, which are erroneous. Notably, the performance gap on entity prediction is huge. These highlight the challenge with the pipeline approach. SpeechRE$_{e2e}$ does not outperform SpeechRE$_{pipe}$ across the three datasets. 

All SpeechRE methods have achieved low accuracy of entity recognition. Particularly, the gap between TextRE and SpeechRE on entity detection is far larger than the gap on relation classification. It suggests that speech entity recognition may be the core bottleneck of the performance degradation of triplet extraction.

\subsection{SpeechRE in Low-resource Scenarios}\label{low}

To evaluate and compare the performance of our SpeechRE models in resource-constrained conditions, we simulate different training conditions by sampling 20\%, 40\%, 60\%, 80\% (and 100\%) from the ReTACRED10 training set. We use REBEL as the TextRE model, since both REBEL and our SpeechRE models are generative models. For evaluation, we randomly sample 20 instances for each relation (10 relations) from the ReTACRED10 test set, totally 200 instances (the same subset corresponding to our human-read test set, described in \S\ref{data}). We present F1 scores on entity prediction in the left plot of Figure \ref{fig:lrda}, and F1 scores on relation prediction in the right plot, in both of which the lines left to the red dashed vertical line refer to the setting discussed in this subsection. 
To measure the extent of error propagation in SpeechRE$_{pipe}$, we also evaluate its performance when the TextRE modules are trained on noisy ASR output as their input (instead of ground-truth transcripts). 
We summarize our observations from different perspectives below.

    \stitle{Training with transcripts v.s. ASR outputs.}
    To investigate the impact of the quality of the text input to REBEL, we compare the performance of REBEL and SpeechRE$_{pipe}$ models, referring to TextRE (Transcript, TTS) and SpeechRE$_{e2e}$ (Single speaker, TTS) in Figure~\ref{fig:lrda}, whose inputs are transcripts and ASR outputs, respectively. Overall, the SpeechRE$_{pipe}$ model, compared to REBEL, produces comparable, yet slightly lower results on relation prediction, whereas performing significantly worse on entity prediction. On the one hand, this indicates the reliability of transcribed texts on relation words. On the other hand, the significant ASR errors on entity words are propagated to the downstream extraction module, greatly degrading its performance. 
    
    \stitle{Pipeline v.s. end-to-end SpeechRE.}
    Comparing the two approaches, SpeechRE$_{e2e}$ performs worse than SpeechRE$_{pipe}$, referring to SpeechRE$_{pipe}$ (ASR, TTS) and SpeechRE$_{e2e}$ (Single speaker,TTS) in Figure~\ref{fig:lrda}. However, with the increase in training data, its performance starts to catch up with SpeechRE$_{pipe}$. This is expected\footnote{The trend has long been observed in other speech processing tasks \cite{sperber2020speech}.}, because not only does SpeechRE$_{pipe}$ have a bigger model size, its two components also excel in their own tasks by leveraging abundant ASR and TextRE data. In comparison, SpeechRE$_{e2e}$ has a smaller model size and is trained with a much smaller training set. Despite the large gap, the rising trend is promising, indicating the potential of  SpeechRE$_{e2e}$ reaching parity with, and even surpassing SpeechRE$_{pipe}$.
    
    \stitle{Training SpeechRE$_{e2e}$ with multi-speaker v.s. single-speaker.} 
     We examine the impacts of single-speaker and multiple-speaker data. 
     In most cases, when a model is trained with multi-speaker data, it has better performance on   relational classification than the one trained on single-speaker data. Their performance on entity recognition is roughly the same.
    
    \stitle{Evaluation on synthetic and human-read data.} 
    When comparing the performance of our models on the synthetic test set and the human-read test set, it is surprising to obverse that most of the time, both SpeechRE$_{pipe}$ and SpeechRE$_{e2e}$ models have higher accuracy on relation prediction on the human-read data than on the synthetic one. This demonstrates the effectiveness of the use of synthetic speech. 
    Please see \emph{Appendix} \ref{appendix:lr} for full results.

\subsection{SpeechRE with Data Augmentation}

\begin{figure*}[t]
    \centering
    \includegraphics[width=\textwidth]{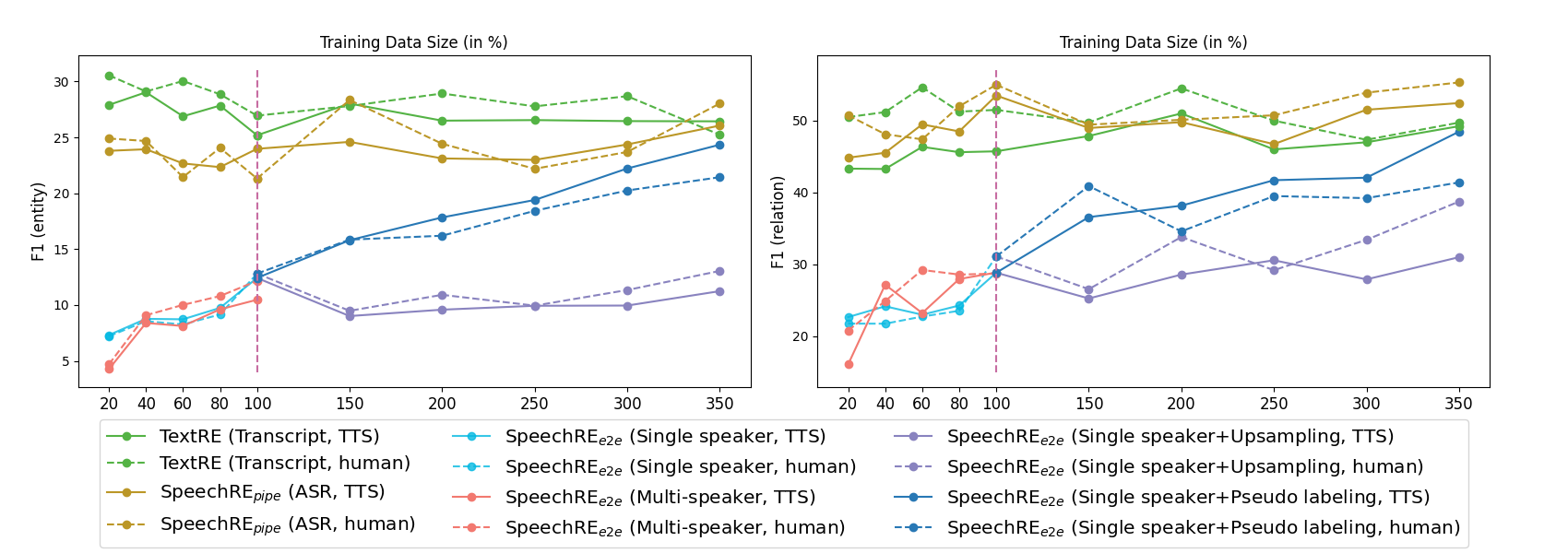}
    \caption{F1 scores of entity (left) and relation (right) predictions on 200 synthetic and human-read instances in various training resource conditions. \textbf{Left to vertical dashed lines}: low-resource scenarios. \textbf{Right to  vertical dashed lines}: data augmentation. \textbf{MODEL} (\textit{Train}, \textit{Test}): MODEL (i.e.\ SpeechRE and TextRE) is trained on \textit{Train} and tested on \textit{Test}.} 
    
    \label{fig:lrda}
\end{figure*}

Based on the trend observed previously,
we expect the SpeechRE$_{e2e}$ model to improve with more training data. We leverage the two data augmentation methods introduced in \S\ref{data_aug}, namely, \textit{up-sampling} and \textit{pseudo-labelling}. For each method, we build larger training corpora by adding augmented SpeechRE data to ReTACRED10, with sizes 100\%, 200\%, 250\%, 300\% and 350\% that of ReTACRED10.  This gives us 10 new
training sets. The evaluation protocol is identical to the one in \S\ref{data_aug}. Results of these data augmentation can be found in Figure ~\ref{fig:lrda}, to the right of the vertical dashed line in each subfigure. We outline our findings below.



    \stitle{Training with transcripts v.s. ASR Outputs.} \
    With more training data, REBEL trained on ground-truth transcripts plus augmented pairs, <\cmss{transcript, pseudo\_triplet}>, has roughly the same accuracy on relation prediction in all conditions. We can observe a slightly decreasing trend on entity prediction. The performance of SpeechRE$_{pipe}$ has a moderately rising trend before leveling out. 
    
    \stitle{Pipeline v.s.\ end-to-end SpeechRE.}
    Both data augmentation techniques bring significant improvements to SpeechRE$_{e2e}$ with \textit{pseudo-labelling} being superior. 
    \textit{Pseudo-labelling} reaches the same performance both on entity and relation predictions as TextRE on synthetic speech at 350\%. The results are surprising, especially with entity generation, considering the difficulty of the task in the speech domain in general.  
    In contrast, augmented data do not help much with SpeechRE$_{pipe}$ due to the error prorogation issue discussed above.  Please see \emph{Appendix} \ref{appendix:da} and \ref{appendix:entity} for full results.
    
    
    

\subsection{Case Study}\label{case_study}

\begin{table*}[]
\resizebox{\textwidth}{!}{
\begin{tabular}{@{}l|l|l|l@{}}
\toprule
\multirow{2}{*}{Text}      & When bin Laden fled the US   invasion in 2001 , he took refuge with \textcolor{blue}{Haqqani} in a safe house        & \multirow{2}{*}{Golden}   & \multirow{2}{*}{<triplet> \textcolor{blue}{Ahmed   Rashid} <subj> Pakistani <obj> person origin} \\
                           & between the Afghan city of \textcolor{blue}{Khost} and \textcolor{blue}{Miran   Shah}, according to Pakistani   author \textcolor{blue}{Ahmed Rashid}. &                           &                                                                                \\ \toprule
\multirow{2}{*}{TTS ASR}   & When bin-laden   fled the U-S invasion   in 2001, he took refuge with \textcolor{red}{Hakone} in a safe house       & \multirow{2}{*}{SpeechRE$_{pipe}$} & \multirow{2}{*}{<triplet> \textcolor{red}{Akmed Rashid} <subj> Pakistani <obj> person origin}   \\
                           & between   the Afghan City of \textcolor{red}{Coast}   and \textcolor{red}{Muran Shaw}, according to Pakistani author \textcolor{red}{Akmed Rashid}.   &                           &                                                                                \\ \toprule
\multirow{2}{*}{Human ASR} & When bin Laden fled the U-S invasion in 2001, he took refuge with \textcolor{red}{Hakwani} in a safe house          & \multirow{2}{*}{SpeechRE$_{pipe}$} & \multirow{2}{*}{<triplet> \textcolor{blue}{Ahmed Rashid} <subj> author <obj> person title}       \\
                           & between the Afghan City of \textcolor{red}{Cost} and \textcolor{red}{Mirishah}, according to Pakistani author \textcolor{blue}{Ahmed Rashid}.          &                           &                                                                                \\ \midrule \midrule
TTS Audio                  & {\href{https://github.com/wutong8023/SpeechRE/blob/master/test-1248_TTS.wav}{Synthetic Audio \faIcon[]{file-audio}}}                                                                                              & SpeechRE$_{e2e}$                   & <triplet>\textcolor{red}{Bernama}<subj>U.S.<obj>organization country of   branch                \\ \midrule
Human Audio                & {\href{https://github.com/wutong8023/SpeechRE/blob/master/test-1248_human.wav}{Human Audio \faIcon[]{file-audio}}}                                                                                              & SpeechRE$_{e2e}$                   & <triplet>\textcolor{red}{Mohamed ElBaradei}<subj>Sultan<obj>person title                        \\ \bottomrule
\end{tabular}
}
\caption{A qualitative error analysis for both the pipeline and end-to-end approaches. Models are trained with 100\% ReTACRED10 data.}
\label{tab:case_study}
\end{table*}
We perform a qualitative error analysis of SpeechRE through a case study. Table~\ref{tab:case_study} shows typical errors in this task. 

    \stitle{Error accumulation in the pipeline method.} 
    The two rows ``TTS ASR'' and ``Human ASR'' illustrate that it is challenging for the SOTA ASR model to spell entity names correctly, especially the names of people and institutions. Being a deep learning model, it may tend to generate high-frequency words~ \cite{razeghi2022impact}.
    This presents both a great challenge and opportunity for entity-sensitive tasks such as relation extraction, since low-frequency entities often contain more information and are more likely to be useful knowledge.
    
    \stitle{Hallucination in the end-to-end method.} 
    As shown in the two rows ``SpeechRE$_{e2e}$'',  the e2e model may generate entities and relation types that are not present in speech, creating hallucinations. TextRE, in contrast, can restrict generated words via controllable text generation techniques. This is less surprising: being a cross-modal task, it is difficult for a SpeechRE model to effectively restrict the generated content, especially when the size of training data is limited. 
We also detail the impact of data augmentation on the accuracy of entity prediction by entity type.  \emph{Appendix} \ref{appendix:entity} contains further results.

\section{Discussion}
Based on our analysis, we discuss the following question: \textbf{For a new SpeechRE task, should we choose a pipeline or an end-to-end approach?}
%
While raw performance is largely attributed to data resources, to answer this question, other factors need to be taken into account in addition to the availability of data resources. These include compute power and latency.

    \stitle{Pipeline method is suitable in the low-resource scenarios.} As shown in Figure \ref{fig:lrda}, prior to 100\%, SpeechRE$_{pipe}$ requires less training data to train than SpeechRE$_{e2e}$, while exhibiting reasonably good performance.
    Therefore, the general ASR method based on pre-training provides a reasonable performance lower bound for low-resource speech extraction. 
    The major concern is errors contained in entities, as shown in \S\ref{case_study}. As a future direction, we conjecture that this issue could be potentially alleviated by the mixed extraction method from both transcript and speech. 
    Yet, the pipeline approach may be limited by fundamental issues (e.g.\ error propagation and high latency) that cannot be solved easily.
    
    \stitle{The end-to-end method is preferred when labelled training data size is sufficient or external data is accessible.} 
    According to Figure \ref{fig:lrda}, with the increasing volumes in training set, the performance of SpeechRE$_{e2e}$ on extracting correct entities and relations steadily improves.
    Extracting meaning from speech directly avoids the risk of information loss and error propagation, unlike in the pipeline setting. Because of this, the e2e approach can potentially solve the extraction task in a principled manner. 
    The data scarcity issue that it faces can be eased through data augmentation, for its effectiveness on both machine-generated speech and realistic speech. Exploring more sophisticated augmentation and filtering techniques is thus a fruitful future direction.
    Further, it is of importance to improve data efficiency and enhance entity prediction performance. Particularly, enforcing constraints on entities in the decoding process and the inclusion of memory banks are promising directions. 

\section{Conclusion}
We propose a new spoken language understanding task, Speech Relation Extraction (SpeechRE), and present two synthetic datasets and a human-read test set. We approach SpeechRE with two methods, the pipeline and e2e approaches. Through extensive experiments, quantitative and qualitative analyses, we identify data scarcity and spoken entity recognition as two main challenges for this task. We then present two augmentation techniques that are effective in addressing these challenges. Lastly, being the first working on the task, we outline key directions for future research.


\section{Limitations}
This paper discusses the utterance-level speech relation extraction task where the average length of audio inputs is less than 15s. Constrained by computing resources, processing long audio signals is challenging, a known issue in the speech domain. For this reason, while speech relations can be useful in other scenarios such as summarization of dialogues, news and meetings, we were not in the position to carry out our study in these scenes. Another limitation is that we did not fully utilize the information contained in speech signals (e.g.\ speaker style and emotion), which could be beneficial to the task. Addressing these two limitations is part of our plan for future research.



\bibliography{anthology,custom}
\bibliographystyle{acl_natbib}

\newpage
\clearpage
\appendix
\section{Appendix}


\subsection{Training Details}\label{appendix:training details}
\subsubsection{Implementation Details}
For all the experiments, we train our REBEL for 30 epochs with Adam optimizer (0.9, 0.999) of a linear scheduler with a warmup rate of 0.1, a learning rate of 5e-5, a weight decay of 0.01, and a gradient clip value of 10. For other settings, our REBEL is consistent with the original ones.

\subsubsection{More details about pseudo-labelling}

The Common Voice Corpus 9.0 dataset consists of 2,224 validated hours in English and 81,085 voices. Each entry in the dataset consists of a unique MP3 and corresponding text file. Many of the recorded hours in the dataset also include demographic metadata like age, sex, and accent that can help train speech recognition engines. Here we use the script from speechbrain\footnote{\url{https://github.com/speechbrain/speechbrain/blob/develop/recipes/CommonVoice/common_voice_prepare.py}} to help us process text files in the dataset, but we made two changes to the processing script. Firstly, this processing script will make all text uppercase, which we do not do, but retain the original case of the text. Secondly, we add full stops to all sentences, whereas the original text has no full stops. We have made these two changes to make the processed text more realistic and to harmonise it with other datasets (e.g. CoNLL04, ReTACRED, etc.). We fine-tune REBEL (using rebel-large\footnote{\url{https://huggingface.co/Babelscape/rebel-large}} as the pretrained model) on the ReTACRED10 dataset and conduct pseudo-labelling on the processed text to extract sentences and corresponding audio that contains target relations. A total of 922k instances were extracted from the Common Voice Corpus 9.0 dataset, of which 380k clean instances were retained after filtering.


\subsection{Low-Resource Analysis}\label{appendix:lr}
\begin{table*}[htb]
\centering
\resizebox{\textwidth}{!}{
\begin{tabular}{@{}|l|l|l|ll|ll|ll|ll|ll|@{}}
\toprule
\multirow{2}{*}{Method}        & \multirow{2}{*}{Input}          & \multirow{2}{*}{Metrics} & \multicolumn{2}{l|}{20\%}            & \multicolumn{2}{l|}{40\%}            & \multicolumn{2}{l|}{60\%}            & \multicolumn{2}{l|}{80\%}            & \multicolumn{2}{l|}{100\%}           \\ \cmidrule(l){4-13} 
                               &                                 &                          & \multicolumn{1}{l|}{w/TTS} & w/Human & \multicolumn{1}{l|}{w/TTS} & w/Human & \multicolumn{1}{l|}{w/TTS} & w/Human & \multicolumn{1}{l|}{w/TTS} & w/Human & \multicolumn{1}{l|}{w/TTS} & w/Human \\ \midrule
\multirow{3}{*}{TextRE}        & \multirow{3}{*}{Transcript}     & Entity                   & \multicolumn{1}{l|}{27.91} & 30.56   & \multicolumn{1}{l|}{29.04} & 29.11   & \multicolumn{1}{l|}{26.89} & 30.05   & \multicolumn{1}{l|}{27.85} & 28.86   & \multicolumn{1}{l|}{25.19} & 26.95   \\ \cmidrule(l){3-13} 
                               &                                 & Relation                 & \multicolumn{1}{l|}{43.32} & 50.50   & \multicolumn{1}{l|}{43.27} & 51.19   & \multicolumn{1}{l|}{46.35} & 54.64   & \multicolumn{1}{l|}{45.61} & 51.26   & \multicolumn{1}{l|}{45.73} & 51.50   \\ \cmidrule(l){3-13} 
                               &                                 & Triplet                  & \multicolumn{1}{l|}{6.55}  & 9.50    & \multicolumn{1}{l|}{6.86}  & 8.45    & \multicolumn{1}{l|}{5.54}  & 8.52    & \multicolumn{1}{l|}{8.02}  & 8.54    & \multicolumn{1}{l|}{4.02}  & 7.50    \\ \midrule
\multirow{3}{*}{SpeechRE-pipe} & \multirow{3}{*}{ASR}            & Entity                   & \multicolumn{1}{l|}{23.80} & 24.90   & \multicolumn{1}{l|}{23.94} & 24.68   & \multicolumn{1}{l|}{22.67} & 21.45   & \multicolumn{1}{l|}{22.34} & 24.06   & \multicolumn{1}{l|}{23.98} & 21.32   \\ \cmidrule(l){3-13} 
                               &                                 & Relation                 & \multicolumn{1}{l|}{44.84} & 50.75   & \multicolumn{1}{l|}{45.53} & 48.11   & \multicolumn{1}{l|}{49.48} & 47.40   & \multicolumn{1}{l|}{48.50} & 52.00   & \multicolumn{1}{l|}{53.50} & 55.00   \\ \cmidrule(l){3-13} 
                               &                                 & Triplet                  & \multicolumn{1}{l|}{4.53}  & 4.02    & \multicolumn{1}{l|}{7.05}  & 5.41    & \multicolumn{1}{l|}{4.64}  & 3.65    & \multicolumn{1}{l|}{4.50}  & 4.50    & \multicolumn{1}{l|}{6.00}  & 4.50    \\ \midrule
\multirow{3}{*}{SpeechRE-e2e}  & \multirow{3}{*}{Single Speaker} & Entity                   & \multicolumn{1}{l|}{7.31}  & 7.2     & \multicolumn{1}{l|}{8.76}  & 8.55    & \multicolumn{1}{l|}{8.72}  & 8.26    & \multicolumn{1}{l|}{9.74}  & 9.15    & \multicolumn{1}{l|}{12.42} & 12.81   \\ \cmidrule(l){3-13} 
                               &                                 & Relation                 & \multicolumn{1}{l|}{22.65} & 21.74   & \multicolumn{1}{l|}{24.15} & 21.73   & \multicolumn{1}{l|}{22.98} & 22.72   & \multicolumn{1}{l|}{24.26} & 23.53   & \multicolumn{1}{l|}{28.83} & 31.09   \\ \cmidrule(l){3-13} 
                               &                                 & Triplet                  & \multicolumn{1}{l|}{0.45}  & 0.45    & \multicolumn{1}{l|}{1.87}  & 0.96    & \multicolumn{1}{l|}{1.9}   & 1.44    & \multicolumn{1}{l|}{2.78}  & 2.34    & \multicolumn{1}{l|}{2.95}  & 2.16    \\ \midrule
\multirow{3}{*}{SpeechRE-e2e}  & \multirow{3}{*}{Multi Speaker}  & Entity                   & \multicolumn{1}{l|}{4.26}  & 4.69    & \multicolumn{1}{l|}{8.39}  & 9.08    & \multicolumn{1}{l|}{8.12}  & 10.01   & \multicolumn{1}{l|}{9.61}  & 10.8    & \multicolumn{1}{l|}{10.47} & 12.16   \\ \cmidrule(l){3-13} 
                               &                                 & Relation                 & \multicolumn{1}{l|}{16.06} & 20.77   & \multicolumn{1}{l|}{27.16} & 24.88   & \multicolumn{1}{l|}{23.21} & 29.21   & \multicolumn{1}{l|}{27.95} & 28.57   & \multicolumn{1}{l|}{28.85} & 28.71   \\ \cmidrule(l){3-13} 
                               &                                 & Triplet                  & \multicolumn{1}{l|}{0.95}  & 0.95    & \multicolumn{1}{l|}{2.35}  & 1.42    & \multicolumn{1}{l|}{1.45}  & 1.91    & \multicolumn{1}{l|}{2.27}  & 2.67    & \multicolumn{1}{l|}{2.27}  & 1.83    \\ \bottomrule
\end{tabular}
}
\caption{The low resource analysis.}
\label{tab:low_resource}
\end{table*}

We report the exact values of low resource analysis in Table~\ref{tab:low_resource}, which corresponds to the left half of each sub-figure of Figure~\ref{fig:lrda}.

\subsection{Data Augmentation Analysis}\label{appendix:da}
\begin{table*}[]
\centering
\resizebox{\textwidth}{!}{
\begin{tabular}{@{}|l|l|l|ll|ll|ll|ll|ll|ll|@{}}
\toprule
\multirow{2}{*}{Method}        & \multirow{2}{*}{Input}                         & \multirow{2}{*}{Metrics} & \multicolumn{2}{l|}{100\%}           & \multicolumn{2}{l|}{100\% + 50\%}    & \multicolumn{2}{l|}{100\% + 100\%}   & \multicolumn{2}{l|}{100\% + 150\%}   & \multicolumn{2}{l|}{100\% + 200\%}   & \multicolumn{2}{l|}{100\% + 250\%}   \\ \cmidrule(l){4-15} 
                               &                                                &                          & \multicolumn{1}{l|}{w/TTS} & w/Human & \multicolumn{1}{l|}{w/TTS} & w/Human & \multicolumn{1}{l|}{w/TTS} & w/Human & \multicolumn{1}{l|}{w/TTS} & w/Human & \multicolumn{1}{l|}{w/TTS} & w/Human & \multicolumn{1}{l|}{w/TTS} & w/Human \\ \midrule
\multirow{3}{*}{TextRE}        & \multirow{3}{*}{Text + Pseudo Labeling}        & Entity                   & \multicolumn{1}{l|}{25.19} & 26.95   & \multicolumn{1}{l|}{28.03} & 27.81   & \multicolumn{1}{l|}{26.50} & 28.93   & \multicolumn{1}{l|}{26.55} & 27.78   & \multicolumn{1}{l|}{26.46} & 28.69   & \multicolumn{1}{l|}{26.44} & 25.24   \\ \cmidrule(l){3-15} 
                               &                                                & Relation                 & \multicolumn{1}{l|}{45.73} & 51.50   & \multicolumn{1}{l|}{47.86} & 49.75   & \multicolumn{1}{l|}{51.00} & 54.50   & \multicolumn{1}{l|}{46.00} & 50.00   & \multicolumn{1}{l|}{46.99} & 47.34   & \multicolumn{1}{l|}{49.23} & 49.73   \\ \cmidrule(l){3-15} 
                               &                                                & Triplet                  & \multicolumn{1}{l|}{4.02}  & 7.50    & \multicolumn{1}{l|}{1.40}  & 8.04    & \multicolumn{1}{l|}{6.50}  & 8.50    & \multicolumn{1}{l|}{5.00}  & 7.50    & \multicolumn{1}{l|}{7.18}  & 8.51    & \multicolumn{1}{l|}{7.42}  & 7.41    \\ \midrule
\multirow{3}{*}{SpeechRE-pipe} & \multirow{3}{*}{Transcript + Pseudo Labeling}  & Entity                   & \multicolumn{1}{l|}{23.98} & 21.32   & \multicolumn{1}{l|}{24.60} & 28.35   & \multicolumn{1}{l|}{23.12} & 24.43   & \multicolumn{1}{l|}{22.99} & 22.19   & \multicolumn{1}{l|}{24.37} & 23.70   & \multicolumn{1}{l|}{26.06} & 28.04   \\ \cmidrule(l){3-15} 
                               &                                                & Relation                 & \multicolumn{1}{l|}{53.50} & 55.00   & \multicolumn{1}{l|}{48.98} & 49.44   & \multicolumn{1}{l|}{49.79} & 50.12   & \multicolumn{1}{l|}{46.73} & 50.75   & \multicolumn{1}{l|}{51.52} & 53.90   & \multicolumn{1}{l|}{52.45} & 55.32   \\ \cmidrule(l){3-15} 
                               &                                                & Triplet                  & \multicolumn{1}{l|}{6.00}  & 4.50    & \multicolumn{1}{l|}{4.67}  & 4.88    & \multicolumn{1}{l|}{6.56}  & 7.01    & \multicolumn{1}{l|}{3.52}  & 3.52    & \multicolumn{1}{l|}{6.06}  & 7.56    & \multicolumn{1}{l|}{6.18}  & 5.25    \\ \midrule
\multirow{3}{*}{SpeechRE-e2e}  & \multirow{3}{*}{One Speaker + Upsampling}      & Entity                   & \multicolumn{1}{l|}{12.42} & 12.81   & \multicolumn{1}{l|}{9.01}  & 9.48    & \multicolumn{1}{l|}{9.58}  & 10.91   & \multicolumn{1}{l|}{9.93}  & 9.94    & \multicolumn{1}{l|}{9.95}  & 11.34   & \multicolumn{1}{l|}{11.24} & 13.05   \\ \cmidrule(l){3-15} 
                               &                                                & Relation                 & \multicolumn{1}{l|}{28.83} & 31.09   & \multicolumn{1}{l|}{25.24} & 26.54   & \multicolumn{1}{l|}{28.57} & 33.8    & \multicolumn{1}{l|}{30.56} & 29.18   & \multicolumn{1}{l|}{27.9}  & 33.41   & \multicolumn{1}{l|}{30.99} & 38.74   \\ \cmidrule(l){3-15} 
                               &                                                & Triplet                  & \multicolumn{1}{l|}{2.95}  & 2.16    & \multicolumn{1}{l|}{2.22}  & 2.65    & \multicolumn{1}{l|}{2.19}  & 2.17    & \multicolumn{1}{l|}{2.14}  & 2.59    & \multicolumn{1}{l|}{1.37}  & 0.91    & \multicolumn{1}{l|}{1.87}  & 1.87    \\ \midrule
\multirow{3}{*}{SpeechRE-e2e}  & \multirow{3}{*}{One Speaker + Pseudo Labeling} & Entity                   & \multicolumn{1}{l|}{12.42} & 12.81   & \multicolumn{1}{l|}{15.8}  & 15.85   & \multicolumn{1}{l|}{17.84} & 16.2    & \multicolumn{1}{l|}{19.4}  & 18.43   & \multicolumn{1}{l|}{22.22} & 20.25   & \multicolumn{1}{l|}{24.35} & 21.44   \\ \cmidrule(l){3-15} 
                               &                                                & Relation                 & \multicolumn{1}{l|}{28.83} & 31.09   & \multicolumn{1}{l|}{36.57} & 40.89   & \multicolumn{1}{l|}{38.16} & 34.59   & \multicolumn{1}{l|}{41.71} & 39.5    & \multicolumn{1}{l|}{42.06} & 39.21   & \multicolumn{1}{l|}{48.47} & 41.4    \\ \cmidrule(l){3-15} 
                               &                                                & Triplet                  & \multicolumn{1}{l|}{2.95}  & 2.16    & \multicolumn{1}{l|}{2.56}  & 1.95    & \multicolumn{1}{l|}{3.67}  & 1.99    & \multicolumn{1}{l|}{5.01}  & 2.49    & \multicolumn{1}{l|}{4.9}   & 4.46    & \multicolumn{1}{l|}{7.22}  & 4.99    \\ \bottomrule
\end{tabular}
}
\caption{The data augmentation analysis.}
\label{tab:data_augmentation}
\end{table*}
We report the exact values of low resource analysis in Table~\ref{tab:low_resource}, which corresponds to the right half of each sub-figure of Figure~\ref{fig:lrda}.

\subsection{Entity Analysis}\label{appendix:entity}

 To further understand why the pseudo labelling can perform better than the multi-speaker up-sampling, we conduct the following analysis experiments. 
 Firstly, we selected high-frequency entities with frequency greater than three from the test set of ReTACRED10. Moreover, we count the frequency of these entities in the training set constructed by two augmentation manners, i.e., pseudo labelling and multi-speaker up-sampling (as shown in Table~\ref{tab:entity_freq}). 
 Then, we counted the classification accuracy of these entities in the test set (in Table~\ref{tab:entity_accruacy}). 
Comparing the two tables by location, we can observe that the method of pseudo labelling can effectively improve the recognition accuracy of the model for unseen entities.

\begin{table*}[t]
    \centering
    \includegraphics[width=\textwidth]{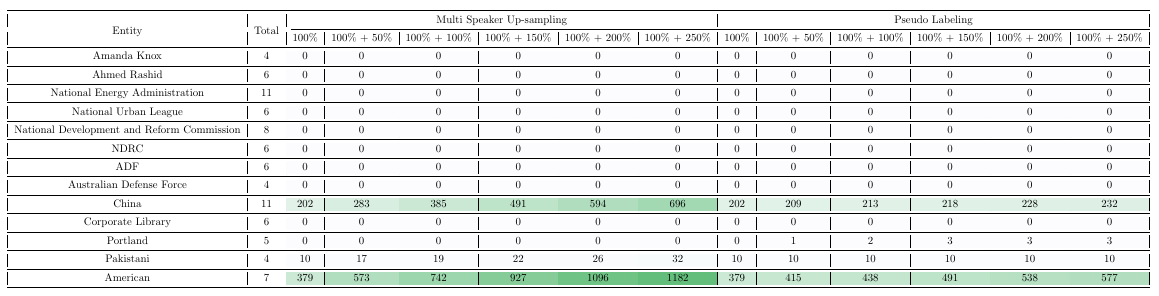}
    \caption{The frequency in the training set of some entities which is demonstrated because their frequency in the test set is greater than 3.}
    \label{tab:entity_freq}
\end{table*}

\begin{table*}[t]
    \centering
    \includegraphics[width=\textwidth]{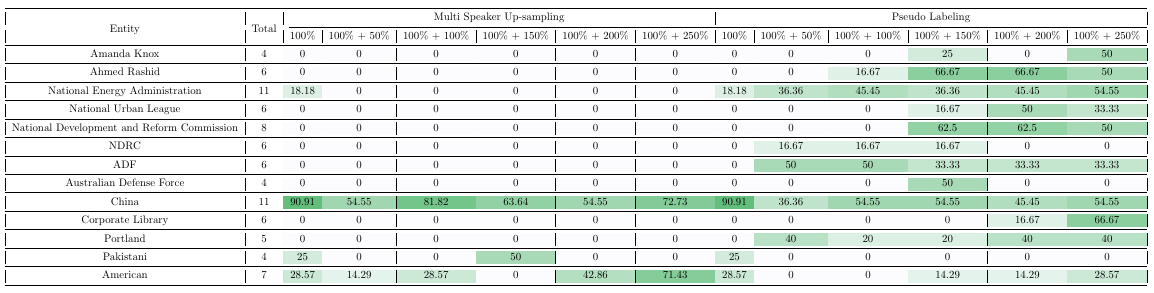}
    \caption{The prediction accuracy of some entities which is demonstrated because their frequency in the test set is greater than 3.}
    \label{tab:entity_accruacy}
\end{table*}

\label{sec:appendix}


\end{document}